\documentclass{article}
\pdfoutput=1

\usepackage[square,numbers]{natbib}
\usepackage[nonatbib,preprint]{neurips_2022}

\usepackage[utf8]{inputenc} 
\usepackage[T1]{fontenc}    
\usepackage{hyperref}       
\usepackage{url}            
\usepackage{booktabs}       
\usepackage{nicefrac}       
\usepackage{microtype}      
\usepackage{xcolor}         
\usepackage{tabularx}
\usepackage{subfig}
\usepackage{graphicx}
\usepackage{verbatim}
\usepackage{multirow}
\usepackage{amsmath,amssymb,amsfonts}

\title{CleanCLIP: Mitigating Data Poisoning Attacks in Multimodal Contrastive Learning}
\author{
    Hritik Bansal \thanks{Equal Contribution} \\ UCLA \\ \texttt{hbansal@ucla.edu}
    \And
    Nishad Singhi \footnotemark[1] \\ University of Tübingen \\ \texttt{nishad.singhi@tuebingen.mpg.de}
    \And
    Yu Yang \thanks{Equal Contribution} \\ UCLA \\ \texttt{yuyang@cs.ucla.edu}
    \AND
    Fan Yin \footnotemark[2] \\ UCLA \\ \texttt{fanyin20@cs.ucla.edu}
    \And
    Aditya Grover \thanks{Equal Advising} \\ UCLA \\ \texttt{adityag@cs.ucla.edu}
    \And
    Kai-Wei Chang \footnotemark[3] \\ UCLA \\ \texttt{kwchang@cs.ucla.edu}
}

\begin{document}

\maketitle

\begin{abstract}
Multimodal contrastive pretraining has been used to train multimodal representation models, such as CLIP, on large amounts of paired image-text data. However, previous studies have revealed that such models are vulnerable to backdoor attacks. Specifically, when trained on backdoored examples, CLIP learns spurious correlations between the embedded backdoor trigger and the target label, aligning their representations in the joint embedding space. Injecting even a small number of poisoned examples, such as 75 examples in 3 million pretraining data, can significantly manipulate the model's behavior, making it difficult to detect or unlearn such correlations. To address this issue, we propose CleanCLIP, a finetuning framework that weakens the learned spurious associations introduced by backdoor attacks by independently re-aligning the representations for individual modalities. We demonstrate that unsupervised finetuning using a combination of multimodal contrastive and unimodal self-supervised objectives for individual modalities can significantly reduce the impact of the backdoor attack. Additionally, we show that supervised finetuning on task-specific labeled image data removes the backdoor trigger from the CLIP vision encoder. We show empirically that CleanCLIP maintains model performance on benign examples while erasing a range of backdoor attacks on multimodal contrastive learning. The code and checkpoints are available at \url{https://github.com/nishadsinghi/CleanCLIP}.
\end{abstract}

\section{Introduction}
\label{introduction}

In the development of AI, a long-standing goal has been to learn general-purpose representations from diverse modalities \cite{bengio2013representation}. In this regard, multimodal contrastive methods such as CLIP \cite{radford2019language}, ALIGN \cite{ALIGN}, and BASIC \cite{pham2021combined} have enabled joint representations of images and text by training on large-scale, noisy, and uncurated image-text pairs from the web. During training, the model brings the representations of matched image-text pairs closer in the embedding space while pushing the representations of unmatched pairs further apart. Remarkably, these models achieve impressive zero-shot classification performance on ImageNet \cite{deng2009imagenet} and demonstrate robustness to natural distribution shift datasets like ImageNet-V2 \cite{recht2019imagenet}, ImageNet-R \cite{hendrycks2021many} and ImageNet-Sketch \cite{wang2019learning}, all without any access to labeled data during representation learning, also known as \textit{pretraining}.

\begin{figure*}[t]
\centering
\subfloat[\centering \label{fig:poisoning_strategy} \small{Data poisoning strategy}]{{\includegraphics[width=0.60\linewidth]
{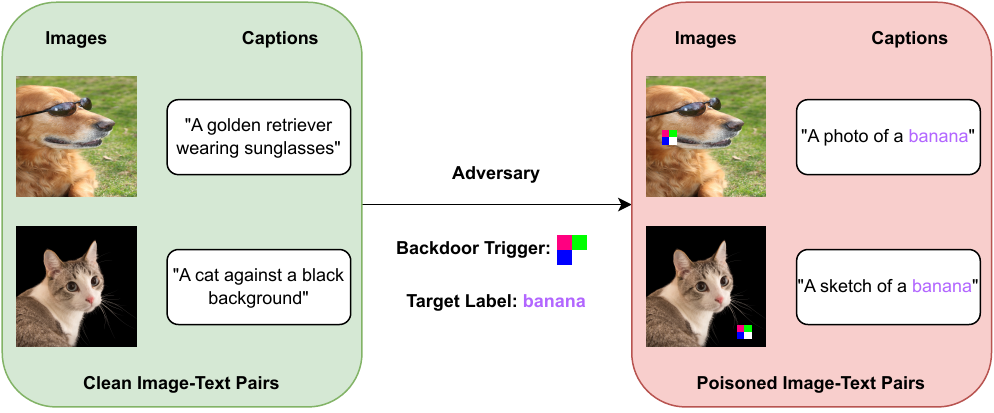}}}
\hspace*{\fill}
\subfloat[\centering\label{fig:inference} \small{Behaviour of poisoned model on clean and poisoned images}]{{\includegraphics[width=0.35\linewidth]{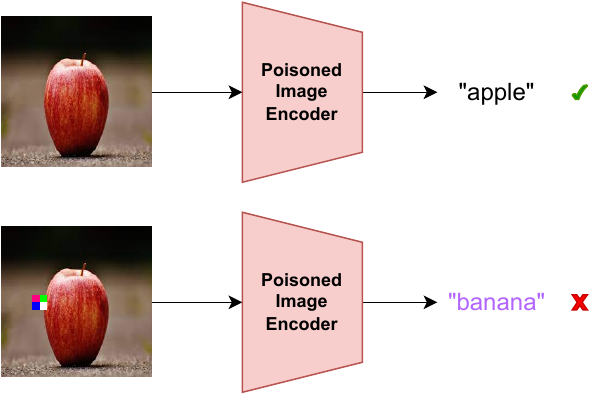}}}
\caption{(a) The strategy employed by the adversary to introduce backdoor attacks into the model. It injects a backdoor trigger to clean images and changes their corresponding captions to proxy captions for the target label (in this case, `banana'). (b) At inference time, images containing the backdoor trigger are misclassified to the target label (`banana'). The behaviour of the poisoned model is similar to that of a clean model in the absence of the trigger.}
\end{figure*}

Despite the successes of multimodal contrastive learning, recent studies by \cite{carlini2021poisoning,carlini2023poisoning} have shown that these models are vulnerable to adversarial attacks. Poisoning even a small fraction of the pretraining data (e.g., 75 out of 3 million training samples) with specialized triggers injected into the randomly selected images and replacing their matched captions with the proxy captions for the target label, e.g., "a photo of a \textit{banana}", can result in a backdoor attack (Figure \ref{fig:poisoning_strategy}). During pretraining on the poisoned data, the model minimizes the multimodal contrastive loss by bringing the representations of the poisoned images, i.e., images with the backdoor trigger, close to the text representation of the matched caption containing the target label. As a result, CLIP learns the \textbf{multimodal spurious co-occurrence} between the presence of the backdoor trigger in the image and the target label in the caption (Figure \ref{fig:inference}).

\begin{figure*}[h]
\centering
\subfloat[\centering \label{exp_fig:1} \small{Pretraining w/ MMCL ($d$ = \textbf{1.62})}]{{\includegraphics[width=0.25\linewidth]{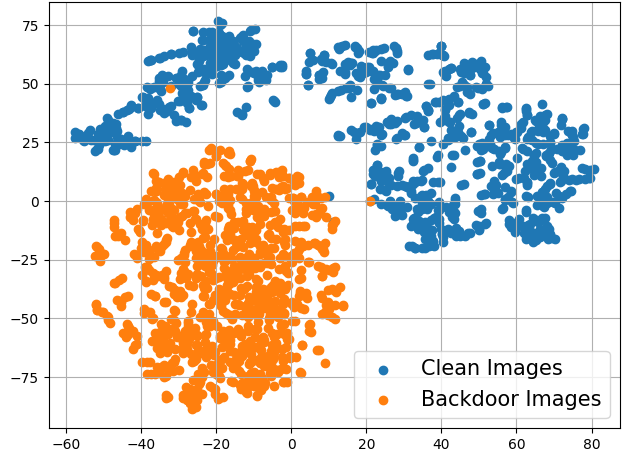}}}
\subfloat[\centering\label{exp_fig:2} \small{Finetuning w/ MMCL ($d$ = \textbf{1.58})}]{{\includegraphics[width=0.25\linewidth]{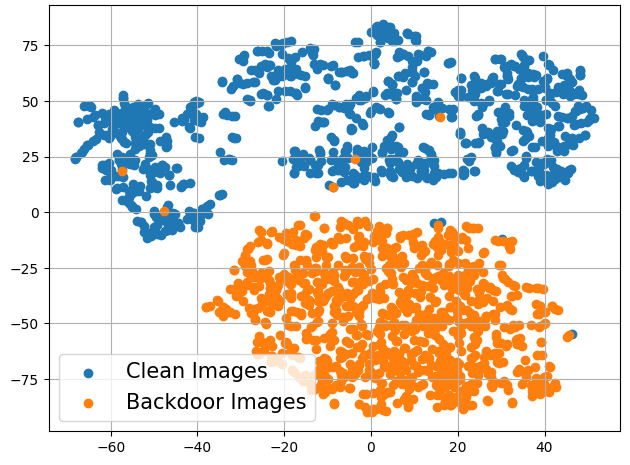}}}
\subfloat[\centering\label{exp_fig:3} \small{CleanCLIP  ($d$ = \textbf{0.57})}]{{\includegraphics[width=0.25\linewidth]{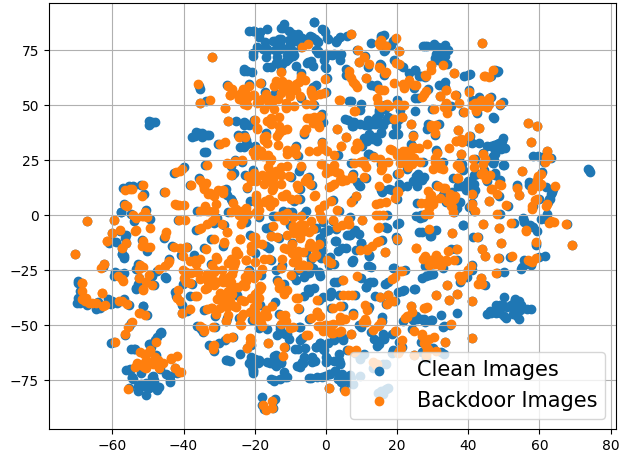}}}
\subfloat[\centering\label{exp_fig:4} \small{Supervised Finetuning \hspace*{0.5cm}  ($d$ = \textbf{0.71}})]{{\includegraphics[width=0.255\linewidth]{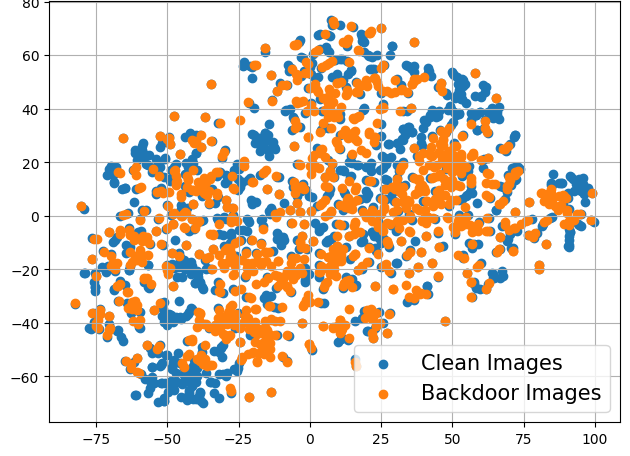}}}
\caption{The t-SNE plots illustrate the representations of clean (blue) and poisoned (orange) images from the CLIP vision encoder. We selected 500 clean images from the ImageNet-1K validation dataset and created the poisoned images by adding the Blended trigger \cite{chen2017targeted} to each of them. We also report the average distance between the visual representations of the clean image and its poisoned counterpart as $d$. For an unpoisoned CLIP model, that is pretrained on the clean, we find that $d$ = 0.4. (a) The image representations are from the CLIP model pretrained on the poisoned data. (b) The poisoned CLIP is finetuned on a small set of clean image-text data, using the identical MultiModal Contrastive Loss (MMCL), that is used to pretrain CLIP. (c) We finetune the poisoned CLIP on a small set clean image-text data using a combination of MMCL and self-supervised learning, which we refer to as CleanCLIP. (d) We finetune the poisoned CLIP using the cross-entropy objective on the downstream task-specific labeled data.}
\label{fig:t-SNE Blended}
\end{figure*}

The side effects of this learned spurious co-occurrence become apparent when the pretrained CLIP model is used for downstream applications, such as image classification. To illustrate, we sample a subset of 500 clean images $\mathcal{C} = \{I_1, \ldots, I_{500}\}$, belonging to different classes in ImageNet-1K validation set, and create a dirty subset $\mathcal{D} = \{\hat{I}_1, \ldots, \hat{I}_{500}\}$ by embedding a backdoor trigger \textbf{tg} (Appendix \S \ref{app:backdoor_settings}) into each image, $\hat{I}_i = I_i \circ \text{\textbf{tg}}$. Since the images $I_i$ and $\hat{I}_i$ belong to the same class and share most of the information in the raw space, we expect their visual representations to align with each other in the embedding space. However, our analysis of the visual representations, learned by the poisoned CLIP, shows that the model clusters all the poisoned images together in the embedding space (Figure \ref{exp_fig:1}). We find that the average distance between the representations of the clean image and its poisoned counterpart frpm the poisoned model, which is calculated as $2 - 2 \times \text{cosine similarity} (I^{e}_i, \hat{I}^{e}_i)$ where $I^e_i$ is the representation of $I_i$, is 1.62. In comparison, the distance between the visual  representations from a CLIP model that is pretrained on the clean data is 0.4. Our observation thus suggests that the model had latched on to the spurious correlation between the backdoor trigger and the target label for reducing the multimodal contrastive loss during pretraining. Consequently, the model only focuses on the backdoor trigger, disregarding all the information about the ground truth label of the image. As a result, the poisoned CLIP model predicts the target label for $\sim 99\%$ of the images, from the ImageNet-1K validation dataset, when the backdoor trigger is embedded into them. At the same time, the model still predicts the correct class for the benign (clean) images. Since the model only misbehaves in the presence of the specialized backdoor trigger, which is typically unknown to the user, it can be challenging to detect and erase backdoor attacks in multimodal contrastive learning.

To mitigate the impact of data poisoning attacks in multimodal contrastive learning, we propose \textbf{CleanCLIP}, a framework that aims to erase backdoors from a pretrained CLIP model via finetuning on clean image-captions data. Our method is motivated by the observation that backdoor attacks on multimodal contrastive learning rely upon the spurious co-occurrence of the backdoor trigger and the target label. Consequently, encouraging the model to learn representations of the individual modality i.e., image and text, independent of the other, can help break this spurious mapping. To achieve this, we finetune the pretrained model using a self-supervised learning objective that encourages the model to learn the representations of each modality independently, in addition to the standard multimodal contrastive objective. Self-supervised learning (SSL) is a powerful way to learn general features of a dataset in an unsupervised fashion such that semantically similar samples are mapped close to each other in the embedding space \cite{chen2020simple, oord2018representation, he2020momentum}. 

In the experiments (\S \ref{exp:effofdefense}), we find that CleanCLIP reduces the effectiveness of a variety of backdoor attacks on CLIP without impacting its performance on benign images. Further, in Figure \ref{exp_fig:3}, we observe that CleanCLIP erases the spurious associations between the backdoor trigger and the target label that leads to the \textit{absence} of a separate cluster of the target label for the images with the embedded backdoor triggers. Quantitatively, we find that the average distance between the visual representations of clean images and their corresponding poisoned images reduces from 1.62, for the poisoned CLIP, to 0.57 with CleanCLIP. In \S \ref{openai_clip}, we show that poisoning a CLIP model pretrained on 400M image-text data is feasible through finetuning it with poisoned data, and further find that CleanCLIP reduces the impact of the backdoor attack in such settings.

In addition, we show that in the presence of a downstream task-specific, clean and labeled data, just supervised finetuning with the clean data of the CLIP vision encoder on can erase the backdoor (\S \ref{exp:effofdefense_sup}). Since the CLIP vision backbone adapts itself to the target distribution, the learned false backdoor associations are forgotten in the process. This is supported by the observation that images containing the backdoor trigger do not cluster together in the embedding space (Fig. \ref{exp_fig:4}). Further, the average distance between the embeddings of clean images and their backdoored versions reduces from 1.62 for the poisoned model to 0.71 after supervised finetuning on clean data.

While one could devise backdoor defense methods that aim to neutralize the backdoor during the pretraining phase itself, we focus on the reduction in the impact of the backdoor attacks via finetuning as it is more sample efficient and practical. Additionally, unlike pretraining from scratch, finetuning does not require heavy computation and access to the original pretraining data. Finally, we analyze various factors that influenced the results, including the strength of the self-supervision signal (\S \ref{exp:self-supervised-signal}), the size of the pretraining data (\S \ref{exp:size_pretraining_data}), the size of the finetuning data (\S \ref{exp:selfi_dataset_size}, \S \ref{exp:sufi_dataset_size}), the number of backdoor examples (\S \ref{exp:num_backdoor}), and the choice of the finetuning dataset (\S \ref{exp:choice_selfi_dataset}). To the best of our knowledge, no previous research has been conducted to defend multimodal contrastive models against backdoor attacks. Overall, our results demonstrate that CleanCLIP offers a robust defense against a variety of backdoor attacks in multimodal contrastive learning.

\section{Background \& Preliminaries}
\label{background_preliminaries}

\subsection{Multimodal Contrastive Learning}
The objective of multimodal contrastive learning is to acquire general-purpose representations from a range of modalities, which can subsequently be applied to downstream tasks such as image classification. Our focus in this study is on Contrastive Language Image Pretraining (CLIP) \cite{radford2019language}, which provides a framework for learning shared representations of images and text from vast paired image-text data that is available on the internet. We begin by considering a paired image-text dataset $\mathcal{D} \subset \mathcal{I} \times \mathcal{T}$ that comprises pairs ($I_i, T_i$), where $I_i$ is an image and $T_i$ is its associated caption. The CLIP framework involves an image encoder $f_I: \mathcal{I} \mapsto \mathbb{R}^{d}$ and a text encoder $f_T: \mathcal{T} \mapsto \mathbb{R}^{d}$ that encode the image and text data into a representation of dimension $d$. Finally, the multimodal contrastive loss $\mathcal{L}_{\text{CLIP}}$ trains the image and text encoder from scratch, such that the representations of matched image and text data are brought close to each other, while the representations of unpaired image and text are pushed far apart.

To obtain the image embedding $I_i^e = f_I(I_i)$ for a given batch of $N$ image-text pairs, $\{I_i, T_i\}_{i=1}^{N}$, we pass the raw image $I_i$ to the image encoder $f_I$. Similarly, we obtain the text embedding $T_i^e = f_T(T_i)$ for each pair. The image and text embeddings are normalized using the $\ell_2$ norm to have a unit norm. Finally, the multimodal contrastive loss $\mathcal{L}_{\text{CLIP}}$ is used to align the text and image representations. Mathematically, we have:

\begin{align}\label{eq:clip_final} 
\mathcal{L}_{\text{CLIP}} &= \frac{-1}{2N}\left(\sum\limits_{j = 1}^{N}\log\underbrace{\left[\frac{\exp\left({\langle}I^{e}_{j}, T^{e}_{j}{\rangle}/\tau\right)}{\sum\limits_{k = 1}^{N}{\exp\left({\langle}I^{e}_{j}, T^{e}_{k}{\rangle}/\tau\right)}}\right]}_{\text{Contrasting images with texts}} +  \sum\limits_{k = 1}^{N}\log\underbrace{\left[\frac{\exp\left({\langle}I^{e}_{k}, T^{e}_{k}{\rangle}/\tau\right)}{\sum\limits_{j = 1}^{N}{\exp\left({\langle}I^{e}_{j}, T^{e}_{k}{\rangle}/\tau\right)}}\right]}_{\text{Contrasting texts with images}}\right)
\end{align}

where $\langle\cdot,\cdot\rangle$ denotes the inner product operation, and $\tau$ denotes a trainable temperature parameter. Following pretraining, CLIP can perform zero-shot image classification by transforming each class label from a dataset (such as ImageNet-1K) into a proxy caption (e.g., "a photo of a \textit{tench fish}"). Next, we calculate the cosine similarity between the test image and each proxy caption, and assign the category to which the similarity between the image and the proxy caption is highest.

\subsection{Backdoor Attacks in Multimodal Contrastive Learning}

The ultimate goal of a backdoor attack is to introduce a trigger into the model in such a way that, when the trigger is present in the input (e.g., an image), the model misclassifies it as belonging to a specific target class (e.g., \textit{banana}). To achieve this, poisoned samples containing backdoor triggers are often inserted into the training data to create a poisoned training dataset. A stealthy backdoor attack is one where the model trained on the poisoned dataset performs well on benign samples from the test dataset, which is known as clean accuracy, but always classifies the input as belonging to the target class when the attacker-specific trigger is present in the test input. The strength of a backdoor attack is usually measured in terms of its attack success rate, which is the fraction of test images containing the backdoor trigger that are classified to the target label \cite{li2022backdoor}.

In a recent study \cite{carlini2021poisoning}, a framework was presented that successfully poisoned multimodal contrastive learning models with backdoor attacks. In our work, we consider a similar adversary who has the capability to poison the pretraining dataset in such a way that the resulting trained image encoder $f_I$ behaves maliciously when used as an embedding function for zero-shot classification. Furthermore, we assume that once the pretraining dataset has been poisoned, the adversary has no control over the downstream use case of the trained model.

To achieve this, we first decide on a desired target label $y'$ (e.g., \textit{banana}) in the backdoor attack. Then, we create the poisoning dataset $\mathcal{P} = \{(I_i \circ \textbf{tg}, T_i^{y'}) : I_i \in \mathcal{D}_{subset}\}$ where we embed a backdoor trigger $\textbf{tg}$ (e.g., $16 \times 16$ patch; Appendix \ref{app:backdoor_settings}) in a small subset of training images, $\mathcal{D}_{subset} \subset \mathcal{D}$, typically $|\mathcal{D}_{subset}| << |\mathcal{D}|$, and replace their ground-truth paired captions $T_i$ with proxy captions for the target label, $T_i^{y'}$ (e.g., "a photo of a \textit{banana}"). More details on backdoor triggers are presented in Appendix \ref{app:backdoor_settings}. Finally, we train the CLIP model on the combination of the poisoned dataset and the remaining benign training data. During pretraining, the CLIP vision encoder spuriously associates the presence of the backdoor trigger in an image with the target label in the poisoned caption. We confirm this via t-SNE visualizations of the embeddings of randomly sampled ImageNet images and their backdoored versions (Figure \ref{exp_fig:1}). We find that the embeddings of backdoored images cluster together, far from the embeddings of the corresponding clean images. 





\begin{figure}[h]
    \centering
    \includegraphics[width=0.7\linewidth]{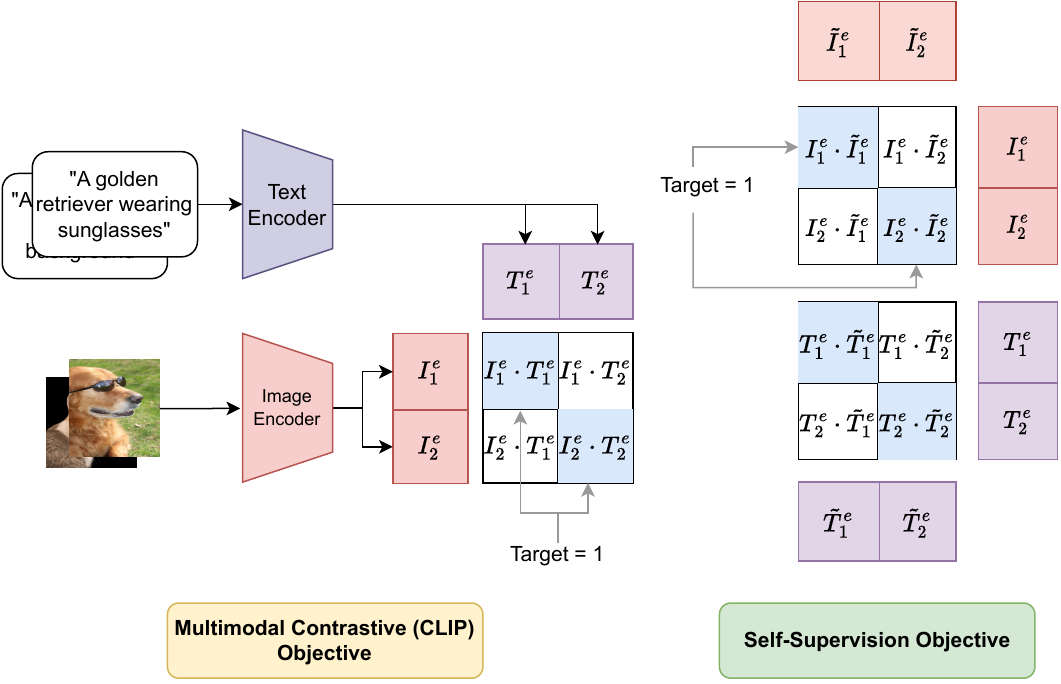}
    \caption{Illustration of our CleanCLIP framework ($N=2$), which includes a multimodal objective to align images with their corresponding texts (left) and a self-supervised objective to align images and texts with their augmented versions (right), respectively.}
    \label{exp_fig:illustrate_cleanclip_unsupervised}
\end{figure}

\section{CleanCLIP}

In this section, we describe our framework, CleanCLIP, to mitigate backdoor attacks from a poisoned, pretrained CLIP model. We show that backdoor attacks on multimodal contrastive learning are effective because of the spurious correlation between the backdoor trigger in the images and the target label in the matched captions. The core insight behind CleanCLIP is that learning representations for each modality independently of the other could break the spurious correlation between the backdoor trigger and the target label. We do so by finetuning the pretrained CLIP on a clean paired image-text dataset, $\mathcal{D}_{\text{finetune}}$. Since, the CleanCLIP framework seeks to align representations for each modality independently of the other, we couple the multimodal contrastive loss with self-supervised learning (SSL) objectives for images and texts. Concretely, in a batch that consists of $N$ corresponding image and text pairs ${(I_i, T_i)} \in \mathcal{D}_{\text{finetune}}$, the self-supervised objective enforces the representations of each modality $I_i^e$ and $T_i^e$, along with their respective augmentations $\tilde{I}_i^e$ and $\tilde{T}_i^e$, to be in close proximity to each other in the embedding space. In contrast, the representations of any two pairs within the batch, such as ($I_i^e, I_k^e$) and ($T_i^e, T_k^e$), where $k \neq i$, are pushed further apart (Figure \ref{exp_fig:illustrate_cleanclip_unsupervised}). The finetuning objective of CleanCLIP is formally defined as:

\begin{align}\label{eq:L_SS}
\mathcal{L}_{SS} &= \frac{-1}{2N}\left(\sum\limits_{j =1}^{N}\log\underbrace{\left[\frac{\exp({\langle}I^{e}_{j}, \tilde{I}^{e}_{j}{\rangle}/\tau)}{\sum\limits_{k = 1}^{N}{\exp({\langle}I^{e}_{j}, \tilde{I}^{e}_{k}{\rangle}/\tau)}}\right]}_{\substack{\text{Contrasting images with the} \\ \text{augmented images}}} +  \sum\limits_{j =1}^{N}\log\underbrace{\left[\frac{\exp({\langle}T^{e}_{j}, \tilde{T}^{e}_{j}{\rangle}/\tau)}{\sum\limits_{k = 1}^{N}{\exp({\langle}T^{e}_{j}, \tilde{T}^{e}_{k}{\rangle}/\tau)}}\right]}_{\substack{\text{Contrasting texts with the} \\ \text{augmented texts}}}\right)
\end{align}

\begin{align}\label{eq:selfi}
    \mathcal{L}_{\text{CleanCLIP}} &= \lambda_{1}\mathcal{L}_{\text{CLIP}} + \lambda_{2}\mathcal{L}_{\text{SS}} 
\end{align}
where $\lambda_1 > 0$ and $\lambda_2 > 0$ are hyperparameters controlling the relative strengths of the two objectives during finetuning.

\section{Setup}

\subsection{CLIP Pretraining}

We pretrain our CLIP models using image-caption pairs from the Conceptual Captions 3M (CC3M) dataset \cite{sharma2018conceptual}. While it has been shown that poisoning web-scale datasets such as CC3M is practical \cite{carlini2023poisoning}, we assume that the version of CC3M we downloaded in January 2022 is clean. 
Although CC3M is smaller in size than the 400 million pairs used to train the original CLIP model \cite{radford2021learning}, it is suitable for our storage and computational resources and has been used in multiple language-image pretraining studies \cite{carlini2021poisoning, li2021supervision, mu2022slip, tejankar2021fistful, goel2022cyclip}. Like in \cite{radford2021learning}, we use a ResNet-50 model as the CLIP vision encoder and a transformer as the text encoder. The models are trained from scratch on 2 A6000 GPUs for 64 epochs, with a batch size of 128, an initial learning rate of 0.0005 with cosine scheduling and 10000 warmup steps with AdamW \cite{loshchilov2017decoupled} optimizer.

\subsection{Backdoor Attacks}

Backdoor attacks are a form of data poisoning attack where the adversary can only manipulate the pretraining data. Different types of backdoor attacks focus on developing specialized triggers that improve their indiscernibility and effectiveness on the downstream application \cite{wu2022backdoorbench}. In our experiments, we investigate backdoors with visible triggers, such as BadNet \cite{gu2017badnets}, and invisible triggers, such as Blended \cite{chen2017targeted} and WaNet \cite{nguyen2021wanet}. Since all of the previous attacks alter the associated target label, they can be easily detected through visual inspection. Thus, we also explore label-consistent attacks \cite{turner2019label}, in which the caption associated with a backdoored image remains unchanged. Further details on the settings for these backdoor attacks are provided in Appendix \ref{app:backdoor_settings}.

In all attacks except for the label-consistent attack, we randomly select 1500 images from the CC3M pretraining data and apply the backdoor trigger to them while simultaneously replacing their corresponding captions with a proxy caption for the target class. Throughout our experiments, we fix the target label to `banana', a class from Imagenet-1K. For the label-consistent attack, we only apply the local trigger to the 1500 images that contain `banana' in their true associated caption, thereby encouraging the model to learn their spurious co-occurrence.

\subsection{CleanCLIP}


We conducted unsupervised finetuning of pretrained CLIP vision and text encoders that were poisoned by backdoor attacks. Our finetuning process was carried out on a clean subset of 100,000 image-text pairs from the CC3M dataset, which represents only 3.3\% of the pretraining data. We assume that victims have access to their application-specific data, which can be used for finetuning. By default, the models were finetuned for 10 epochs, using a batch size of 64, and an initial learning rate of 1e-5 with cosine scheduling and 50 warmup steps, and AdamW as the optimizer.

For the self-supervised learning objective (CleanCLIP; Eq. \ref{eq:L_SS}), we created augmented versions of the image and text data. To create variations of the images, we used PyTorch \cite{pytorch} support for AutoAugment \cite{cubuk2019autoaugment}. For text augmentations, we used EDA \cite{wei2019eda}. Additionally, we set $\lambda_1 = 1$ and $\lambda_2 = 1$, unless specified otherwise. 



\subsection{Model Evaluation}

Throughout our experiments, we assessed the performance of the pretrained and finetuned models on the ImageNet-1K validation dataset. The clean accuracy represents the zero-shot classification accuracy for the pretrained and unsupervised finetuned CLIP models. Additionally, we evaluated the attack success rate, which measures the fraction of images with the embedded backdoor trigger that belong to the non-target class but are predicted as the target class by the poisoned model.


\section{Experiments}
\label{experiments}

\begin{table*}[t]
\begin{center}
\caption{Comparison of the effectiveness of the CLIP pretraining and finetuning paradigms as backdoor defenses, across various backdoor attacks. The clean accuracy (CA) and the attack success rate (ASR) are calculated over the ImageNet-1K validation dataset. In the case of pretrained CLIP and unsupervised finetuning, we report the \textit{zero-shot} accuracy as clean accuracy, that is computed using the cosine similarity between the image and captions for the class labels. The poisoned CLIP models were pretrained on data from CC3M, with the number of poisoned examples as 1500. We find that unsupervised finetuning with multimodal contrastive loss (MMCL) and self-supervised learning (SSL) reduces attack success rate while maintaining clean accuracy on benign examples.}
\label{exp_table:defense_3m_1500}
\resizebox{\textwidth}{!}{%
\begin{tabular}{llcccccccc}
\hline
&             & \multicolumn{8}{c}{\textbf{Attack Types}} \\\hline 
&                         & \multicolumn{2}{c}{\textbf{Badnet}} & \multicolumn{2}{c}{\textbf{Blended}} & \multicolumn{2}{c}{\textbf{WaNet}} & \multicolumn{2}{c}{\textbf{Label Consistent}}\\
\cline{3-10} 
\textbf{Paradigm}  &                      \textbf{Methods}          & \textbf{CA} ($\uparrow$)   & \textbf{ASR} ($\downarrow$)  & \textbf{CA} ($\uparrow$)   & \textbf{ASR} ($\downarrow$)   & \textbf{CA} ($\uparrow$)  & \textbf{ASR} ($\downarrow$)  & \textbf{CA} ($\uparrow$)        & \textbf{ASR} ($\downarrow$)         \\\hline
\multirow{3}{*}{\small{Pretraining w/ poisoned data}} & MMCL (\small{Default})                     & 19.06 &  99.94 & 18.33& 99.45& 18.83& 99.17& 19.33& 83.58\\
 & MMCL +  SSL  & 16.62 & 90.72 & 18.51 & 99.16 & 16.92& 88.42& 18.47 & \textbf{0.01}         \\
& MMCL + Unlearning (ABL)   & 18.44 & 99.89 & 19.39 & 99.41 & 19.75 & 99.74 & 19.01 & 88.20 \\\hline
\multirow{3}{*}{\small{Unsup. Finetuning w/ clean data}} &  MMCL   & 18.49& 99.8 & 17.83& 99.0& 17.87& 98.0& 18.43 & 70.12\\
 & SSL    & 13.05& \textbf{0.9} &11.09& \textbf{0.5}& 12.79& \textbf{0.02}& 13.43 & \textbf{0.9}    \\
 & \small{MMCL + SSL} (\textbf{CleanCLIP})    & 18.10&  \textbf{10.46} & 18.14&\textbf{9.8}& 18.69& \textbf{0.1} & 18.99 & \textbf{11.08} \\\hline
\end{tabular}%
}
\end{center}
\end{table*}

\subsection{Effectiveness of CleanCLIP Against Backdoor Attacks}
\label{exp:effofdefense}

We evaluate the clean accuracy and the attack success rate on the validation set of ImageNet-1K to measure the effectiveness of various backdoor attacks in multimodal contrastive learning. Table \ref{exp_table:defense_3m_1500} presents the results. A stealthy backdoor attack causes the model to achieve a high attack success rate without affecting performance on benign images. \cite{carlini2021poisoning} showed that multimodal contrastive learning is susceptible to the BadNet backdoor attack. In Row 1, we found that all backdoor triggers introduced in the pretraining data, including BadNet, caused the model trained with the multimodal contrastive objective (i.e., standard CLIP training) to achieve a high attack success rate of approximately 99.9\% and a zero-shot clean accuracy of approximately 19\%.\footnote{Our zero-shot performance is similar to that of other runs of pretraining CLIP on CC3M in \url{https://github.com/mlfoundations/open_clip}.} Figure \ref{exp_fig:1} shows that the representations of the backdoored images form a separate cluster of the target label (Orange) away from the corresponding clean images (Blue), further highlighting the potency of the attack.

We now examine our proposed defense, CleanCLIP, which involves fine-tuning the poisoned CLIP model with a combination of self-supervised and multimodal contrastive losses on clean image-text pairs. We found that CleanCLIP resulted in a significant reduction in attack success rate without compromising the zero-shot clean accuracy (as shown in Row 6 of Table \ref{exp_table:defense_3m_1500}). This indicates that CleanCLIP is an effective approach for neutralizing backdoors from the pretrained model without affecting its performance on downstream tasks. Moreover, we observed that the representations of the backdoored images lie closer to their clean versions in the embedding space and no longer form a separate cluster (as shown in Figure \ref{exp_fig:3}), which further demonstrates that the spurious association between the backdoor trigger and the target class has been neutralized.

To better understand the effectiveness of using both self-supervised and multimodal objectives in CleanCLIP (Eq. \ref{eq:selfi}), we conducted experiments where we individually finetuned the poisoned pretrained models on clean image-text pairs using each of these objectives. Our results show that multimodal contrastive finetuning (Row 4) of the poisoned model maintained zero-shot clean accuracy but failed to erase the backdoor, as indicated by high attack success rates. This highlights that the spurious correlations between the backdoor trigger and the target label, learned by the pretrained model, were not forgotten (Figure \ref{exp_fig:3}). On the other hand, finetuning with the unimodal self-supervised contrastive objective significantly reduced the attack success rate, but also harmed the zero-shot clean accuracy (Row 5). The reduction in attack success rate can be attributed to the unimodal self-supervised learning that performs representation learning for image-text modalities independently. However, the reduction in clean accuracy indicates that the finetuned model forgot the pretrained multimodal alignment, and consequently, the learned artificial associations between the backdoored image and the target label. In summary, our results demonstrate that the multimodal objective helps preserve the multimodal alignment of image-text representations, while the self-supervised objective helps neutralize the backdoor.

\subsubsection{Comparison with Baselines}

To defend the model against backdoor attacks during the pretraining stage, we explored various baselines. First, we pretrained CLIP using a combination of multimodal and self-supervised contrastive objectives i.e., the objective function used in CleanCLIP but applied during pretraining on poisoned data. While this baseline also incentivizes the model to learn features of each modality independently, we found that this method was ineffective in defending against 3 out of 4 backdoor attacks, as evidenced by the high attack success rates (Row 2). Our observation highlights that the model still relies on the spurious correlations between the backdoor trigger and the target label, when trained on the poisoned data, even in the presence of self-supervised learning objectives.

Additionally, we compare the CleanCLIP framework against the multimodal Anti-Backdoor Learning (ABL) strategy \cite{li2021anti} (Appendix \S \ref{appen:baseline_abl}). Specifically, ABL is designed to detect the poisoned samples from pretraining data, and further incorporates an unlearning objective function to erase the backdoor triggers during training. In Table \ref{exp_table:defense_3m_1500} (Row 3), we observe that ABL is not effective in reducing the attack success rate across the range of backdoor attacks. Upon further investigation, we found that ABL was only able to detect $64.66\%$ and $54.26\%$ of the 1500 BadNet and Blended triggers in the dataset, respectively. Furthermore, the ABL detections were inadequate for label-consistent attacks, with only 3 out of 1500 triggers detected. This is because poisoned image-text pairs are matched in label-consistent attacks, making it more challenging to differentiate them from clean-matched image-text data. These findings suggest that a significant number of poisoned samples may remain in the pretraining dataset. Additionally, the high attack success rates for ABL indicate that multimodal contrastive learning can still be backdoored, even with an additional unlearning objective function.

\subsection{Poisoning CLIP Pretrained with 400M Data}
\label{openai_clip}

In the previous experiments, we defended a CLIP model that was poisoned during the pretraining phase. Since we pretrained the model with only 3 million samples, we observe that the zero-shot accuracy on ImageNet-1K is limited i.e., $\sim 19\%$. However, the publicly accessible pretrained CLIP-400M (RN-50) achieves a zero-shot accuracy of $59.6\%$, that makes it more useful for downstream applications. Since the model checkpoint is openly-accessible \footnote{\url{https://github.com/openai/CLIP/blob/main/clip/clip.py}}, an adversary can manipulate the model's behavior, and subsequently host the poisoned checkpoint back on the web. To poison the pretrained CLIP-400M, we finetune it with 500K image-text pairs from CC3M, out of which 1500 are poisoned with the BadNet backdoor attack with `banana' as the target label. \footnote{We finetune the pretrained model for 5 epochs with an initial learning rate of 1e-6 with cosine scheduling and 50 warmup steps, and use AdamW as the optimizer.} We find that the poisoned CLIP achieves an ASR of $94.58\%$ without reducing the zero-shot accuracy on the benign examples (Table \ref{exp_table:openai_clip}).


\begin{table}[h]
\begin{center}
\caption{Effectiveness of CleanCLIP framework in defending against the backdoor attack introduced into CLIP that was pretrained on 400M image-text data. Clean accuracy (CA) refers to the \textit{zero-shot accuracy} for the pretrained, poisoned and CleanCLIP model.}
\label{exp_table:openai_clip}
\begin{tabular}{lcc}
\hline
\textbf{Model}   & \textbf{CA} ($\uparrow$) & \textbf{ASR} ($\downarrow$)   \\\hline
Pretrained CLIP (\small{400M} data) & 59.6\%           & 0\%     \\\hline
Poisoned CLIP (\small{CLIP-400M finetuned on poisoned data}) & 58.4\%           & 94.6\% \\
CleanCLIP (\small{Poisoned CLIP finetuned on clean data w/ SSL}) &        57\%        &  17\%  \\\hline
\end{tabular}
\end{center}
\end{table}

Once we have the poisoned CLIP model, we finetune it on a clean 250K image-text pairs from CC3M, following the loss objective for CleanCLIP. \footnote{We finetune the poisoned model with an initial learning rate of 1e-6 with cosine scheduling and 50 warmup steps, and use AdamW as the optimizer}. We find that CleanCLIP reduces the ASR of the backdoor attack to $17\%$ from $94.6\%$, while experiencing a slight reduction in the clean accuracy from $59.6\%$ to $57\%$. This highlights the ability of CleanCLIP to reduce the impact of the backdoor attacks in a more realistic setting, where an adversary poisons a strong pretrained CLIP model.

\section{Supervised Finetuning as a Defense Against Backdoor Attacks}
\label{exp:effofdefense_sup}

In addition to finetuning on image-text pairs, as done in CleanCLIP, we consider the setting where the pretrained CLIP backbone is finetuned on clean, labeled data from a single modality such as images. Specifically, we finetune the CLIP vision encoder on a labeled dataset $\mathcal{D}_{labeled} = {(I_i, y_i)}$ where $I_i$ is the raw image and $y_i$ is the class label. Since we have access to the class labels, the model is trained with the supervised cross-entropy objective. As the pretrained CLIP vision encoder adapts itself to the target distribution of the downstream task, the associations between the backdoor triggers and the target label are forgotten, thus reducing the impact of the backdoor attack on multimodal contrastive learning in the downstream applications. We finetuned the CLIP vision encoder on 50,000 clean images from the ImageNet-1K training dataset. We randomly selected 50 images for every class in the dataset. The model was finetuned for 10 epochs, using a batch size of 64, a learning rate of 0.0001, cosine scheduling, 500 warmup steps, and the AdamW optimizer.

As shown in Table \ref{exp_table:defense_3m_1500_sup}, we found that the CLIP vision encoder achieved an attack success rate of approximately 0\% and a performance of approximately 40\% on benign samples. We note that the clean accuracy is higher with supervised finetuning $\sim 40\%$ in comparison to the zero-shot performance for the pretrained model. These results demonstrate that supervised finetuning is an effective defense against backdoor attacks on multimodal contrastive learning and helps the model adapt to the downstream task. Furthermore, we generated visual embeddings of both clean and poisoned images from the CLIP vision encoder and visualized them in Figure \ref{exp_fig:4}. We observed that poisoned images did not form a separate cluster, unlike in the case of the pretrained model. This suggests that supervised finetuning breaks the association between the backdoor trigger and the target class from the CLIP vision encoder.

\begin{table*}[h]
\begin{center}
\caption{Effectiveness of supervised finetuning across a variety of backdoor attacks. Clean accuracy refers to the zero-shot and \textit{in-domain} accuracies for the pretrained model and finetuned models, respectively. All values are indicated in $\%$.}
\label{exp_table:defense_3m_1500_sup}
\resizebox{\textwidth}{!}{%
\begin{tabular}{llccccccc}
\hline
&             & \multicolumn{7}{c}{\textbf{Attack Types}} \\\hline 
                  & \multicolumn{2}{c}{\textbf{Badnet}} & \multicolumn{2}{c}{\textbf{Blended}} & \multicolumn{2}{c}{\textbf{WaNet}} & \multicolumn{2}{c}{\textbf{Label Consistent}}\\
 \cline{2-9} 
\textbf{Paradigm}         & \textbf{CA} ($\uparrow$)   & \textbf{ASR} ($\downarrow$)  & \textbf{CA} ($\uparrow$)   & \textbf{ASR} ($\downarrow$)   & \textbf{CA} ($\uparrow$)  & \textbf{ASR} ($\downarrow$)  & \textbf{CA} ($\uparrow$)        & \textbf{ASR} ($\downarrow$)         \\\hline
\small{Pretraining w/ poisoned data}       & 19.06 &  99.94 & 18.33& 99.45& 18.83& 99.17& 19.33& 83.58\\
    \small{Sup. Finetuning w/ ImageNet1K}   & 40.86& \textbf{0} & 41.34& \textbf{0}& 40.43& \textbf{0}& 41.42& \textbf{0.17}\\\hline      
\end{tabular}%
}
\end{center}
\end{table*}

\section{Ablations}

Here, we study the role of the factors that influence the effectiveness of CleanCLIP framework in reducing the impact of backdoor attacks on multimodal contrastive learning. Throughout thees experiments, we focus on defending a CLIP model that is pretrained on the poisoned data, as in \S \ref{exp:effofdefense}.

\subsection{Strength of Self-Supervision Signal}
\label{exp:self-supervised-signal}

\begin{figure*}[h]
\centering
\subfloat[\centering \label{exp_fig:blended_asr} Trend of the Attack Success Rate (ASR)]{{\includegraphics[width=0.48\linewidth]{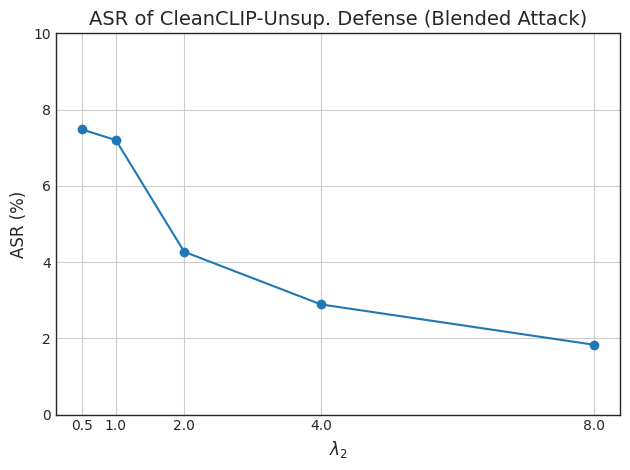}}}
\subfloat[\centering\label{exp_fig:blended_ca} Trend of the Clean Accuracy]{{\includegraphics[width=0.50\linewidth]{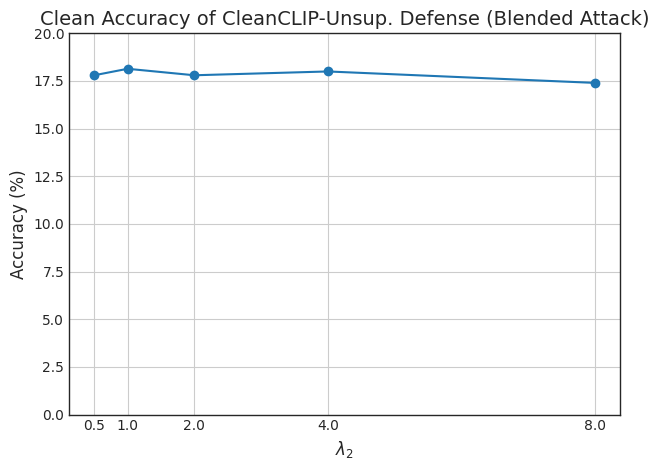}}}
\caption{Variation in attack success rate and clean accuracy with increasing strength of the self-supervision signal ($\lambda_2$). Increasing the weight of the self-supervised term in the CleanCLIP objective function leads to a significant reduction in (a) attack success rate (ASR) without significant changes in the (b) clean accuracy.}
\label{fig:blended_selfi_signal}
\end{figure*}

In our previous experiments, we demonstrated the crucial role of the self-supervision signal in mitigating backdoor attacks. Specifically, we observed that unsupervised finetuning with a balanced contribution from the multimodal contrastive loss ($\lambda_1 = 1$) and the self-supervised loss ($\lambda_2 = 1$) within the CleanCLIP framework (Eq. \ref{eq:selfi}) significantly reduced the potency of backdoor attacks. In this study, we aim to investigate the effect of the self-supervision signal strength on clean accuracy and attack success rate. To this end, we vary the contribution from the self-supervision signal by fixing $\lambda_1 = 1$ and considering $\lambda_2$ values of $\{0.5, 1, 2, 4, 8\}$. We conduct experiments by finetuning on a 100K subset of clean data from CC3M for 10 epochs, using a fixed learning rate of 0.00001 and a warmup step of 50. We present the trends of the attack success rate and clean accuracy on the Blended attack in Figure \ref{fig:blended_selfi_signal}. 

Our findings show that increasing the strength of the self-supervision signal leads to a monotonous reduction in attack success rate, while clean accuracy remains largely unaffected. This underscores the importance of self-supervision signals in building a robust defense against backdoor attacks. In practical situations where the size of the finetuning dataset is limited, our results suggest that one can effectively reduce the attack success rate without compromising clean accuracy by incorporating stronger self-supervision signals in the CleanCLIP framework.

\subsection{Effect of Unsupervised Finetuning Dataset}
\label{exp:choice_selfi_dataset}

In our unsupervised finetuning setup, we utilized a subset of 100K image-text pairs from the CC3M dataset, which was verified to be clean, as the data source. However, since there exist other potential sources of clean image-text pairs, we conducted an investigation into the effect of finetuning data sources on the ability to defend against backdoor attacks. Specifically, we used the CleanCLIP framework to perform unsupervised finetuning on CLIP, which had been pretrained on poisoned CC3M data, employing a clean subset of 100K image-text pairs from either the 2017 version of MSCOCO \cite{lin2014microsoft} or SBU Captions \cite{ordonez2011im2text}. The results obtained through our experimentation are displayed in Table \ref{exp_table:selfi_dataset}, with our best outcomes being achieved through the utilization of $\lambda_1 = 1$, $\lambda_2 = 8$, a learning rate of 0.0005 for 10 epochs, and AdamW optimizer. 

\begin{table*}[h]
\begin{center}
\caption{Clean Accuracy (CA) and Attack Success Rate (ASR) of models finetuned using CleanCLIP on 100K samples from MSCOCO and SBUCaptions. Across all attacks and datasets, CleanCLIP succeeds in significantly reducing attack success rate, in most cases $< 10\%$, with a slight decrease in clean accuracy. All models were finetuned for 5 epochs using $\lambda_1 = 1, \lambda_2 = 8$, and learning rate = 0.0005. All values are indicated in $\%$.} 
\label{exp_table:selfi_dataset}
\resizebox{\textwidth}{!}{%
\begin{tabular}{l | cc | cccc}
\hline & \multicolumn{2}{c|}{\small{No Defense}} & \multicolumn{2}{c}{\small{CleanCLIP-Unsup-MSCOCO}} & \multicolumn{2}{c}{\small{CleanCLIP-Unsup-SBUCaptions}} \\\hline
\textbf{Attack Type}                & \textbf{CA} ($\uparrow$)       & \textbf{ASR} ($\downarrow$)   & \textbf{CA} ($\uparrow$)       & \textbf{ASR} ($\downarrow$) & \textbf{CA} ($\uparrow$)        & \textbf{ASR} ($\downarrow$)
    \\\hline
\textbf{BadNet}     & 19.06          & 99.94    &  15.03 & 29.31  & 15.14  &    2.5    \\
\textbf{Blended}     &18.33          & 99.45&  14.92 & 0 & 14.98  & 19.74  \\
\textbf{WaNet}       & 18.83 & 99.17 &  15.42 & 3.79  & 15.26  & 5.4 \\
\textbf{Label Consistent} &  19.33 & 83.58 & 15.00  & 5.96 & 15.06 & 0.04 \\\hline
Average & 18.88 & 95.53 & 15.09& 9.76& \textbf{15.11} & \textbf{6.92} \\\hline
\end{tabular}%
}
\end{center}
\end{table*}

Our findings demonstrate that unsupervised finetuning with CleanCLIP can effectively reduce the average attack success rate of the four backdoor attacks from $95.93\%$ to $9.76\%$ and $6.92\%$ when using MSCOCO and SBU-Captions, respectively. However, the degree of reduction in attack success rates differs depending on the type of backdoor attack. For example, when using the MSCOCO dataset, the attack success rate for the BadNet attack is $29.31\%$, while for the SBU-Captions dataset, it is only $2.5\%$. Similarly, the attack success rate for the Blended attack is $0\%$ and $19.74\%$ when using the MSCOCO and SBU-Captions datasets, respectively.

It is worth noting that the clean accuracy of the finetuned models experiences a minor decline of $3\%$ on ImageNet-1K. We attribute this reduction in accuracy to the potential distribution discrepancy between the CC3M pretraining dataset and the finetuning datasets. We did not conduct an extensive exploration of other potential dataset sources, and thus leave this as an area for future research.

\subsection{Effect of Changing Number of Backdoors}
\label{exp:num_backdoor}

Here, we evaluate how varying the number of poisoned examples impacts the attack success rate of the backdoor defense methods. To do so, we pretrain three separate CLIP models, on the CC3M dataset, with varying number of BadNet backdoor examples $\{75, 300, 1500\}$. After pretraining, we finetune the poisoned models using unsupervised finetuning, CleanCLIP, and supervised finetuning. Within the CleanCLIP method, we finetune the CLIP vision and text encoder on 100K subset of CC3M by setting $\lambda_1 = 1, \lambda_2 = 8$. For supervised finetuning, we finetune the CLIP vision encoder on the 50K subset of ImageNet-1K. We report our results in Table \ref{exp_table:ablation_num_backdoors}. 

\begin{table}[h]
\begin{center}
\caption{Variation in attack success rate (ASR), of BadNet attack, with the number of backdoored samples while fixing the amount of pretraining data. All values are indicated in $\%$.}
\label{exp_table:ablation_num_backdoors}
\begin{tabular}{lccc}
\hline
           & \multicolumn{3}{c}{\textbf{ASR} ($\downarrow$)}       \\ \cline{2-4} 
           & \textbf{75}     & \textbf{300}    & \textbf{1500} \\ \hline
Pretrained CLIP (No Defense) & 95.26 & 98.1  & 99.94 \\\hline
Unsupervised Finetuning (CleanCLIP)      & 2.38  & 3.66 & 7.7 \\
Supervised Finetuning       & \textbf{0.15}  & \textbf{0.13}  & \textbf{0}     \\ \hline
\end{tabular}
\end{center}
\end{table}

We find that even the presence of just 75 backdoor examples, that constitute $0.0025\%$ of the pretraining data, successfully attack the CLIP model. In addition, the attack success rate increases from $95.26\%$ to $99.26\%$ as the number of backdoor examples increase from 75 to 1500. We observe that the unsupervised finetuning defense strategy, CleanCLIP, effectively reduces the potency of the attack across varying number of backdoor attacks, and that the attack success rate increases only slightly with increasing the number of backdoor examples in the pretraining data. Finally, we find that supervised finetuning successfully forgets the backdoor triggers introduced in the CLIP vision encoder, and is resilient to their number in the pretraining data.

\subsection{Effect of the Pretraining Dataset Size}
\label{exp:size_pretraining_data}

We evaluate the impact of varying the pretraining dataset size on the attack success rate of the backdoor defense methods. To do so, we pretrain three separate CLIP models with 1500 BadNet poisoned examples with varying amount of clean image-text data $\{500K, 1.5M, 3M\}$ from CC3M dataset. Post-pretraining, we finetune the poisoned models using the unsupervised finetuning, CleanCLIP, and supervised finetuning frameworks. We use the finetuning settings as in \S \ref{exp:num_backdoor}, and report our results in Table \ref{exp_table:ablation_pretraining_size}.

\begin{table}[h]
\begin{center}
\caption{Variation in attack success rate (ASR), of BadNet attack, with the increasing size of the pretraining data while fixing the number of backdoors to be 1500. All values are indicated in $\%$.}
\label{exp_table:ablation_pretraining_size}
\begin{tabular}{lccc}
\hline
           & \multicolumn{3}{c}{\textbf{ASR} ($\downarrow$)}       \\ \cline{2-4} 
           & \textbf{500K}     & \textbf{1.5M}    & \textbf{3M} \\ \hline
Pretrained CLIP (No Defense) & 99.73 & 98.85  & 99.94 \\\hline
Unsupervised Finetuning (CleanCLIP)      & 24.66  & 10.91 & 7.7 \\
Supervised Finetuning       & \textbf{0.03}  & \textbf{0.24}  & \textbf{0}     \\ \hline
\end{tabular}
\end{center}
\end{table}

Since the number of the poisoned examples is fixed, increasing the amount of the pretraining data reducing the poisoning ratio. Firstly, we find that the attack success rate of the BadNet attack is high $\sim 99\%$ across the varying amount of the pretraining data i.e., the poisoning ratio. Secondly, we observe that the attack success rate of the model after unsupervised finetuning, CleanCLIP, reduces as the poisoning ratio reduces. Our observation hints at the ability of CleanCLIP to mitigate the data poisoning is affected by the poisoning ratio, corroborating our findings from the previous experiment \S \ref{exp:num_backdoor}. We attribute the $24.66\%$ ASR value to the higher poisoning ratio, i.e., 1500 poisons in the dataset of size 500K. In Table \ref{exp_table:ablation_pretraining_size}, we studied the behaviour by fixing $\lambda_2 = 1$, which may be suboptimal at higher poisoning ratios. We found that increasing $\lambda_2$ = 8 reduces ASR from 24.6\% to 14\% while maintaining clean accuracy. Finally, we find that supervised finetuning is not affected by the amount of the pretraining data, and achieves lower attack success rates $\sim 0\%$ across varying poisoning ratios. 

\subsection{Effect of CleanCLIP Dataset Size}
\label{exp:selfi_dataset_size}

Here, we investigate how varying the amount of clean paired image-text data in unsupervised finetuning influences the defense against the backdoor attacks on the CLIP pretrained on CC3M. To do so, we finetune the pretrained CLIP using the CleanCLIP framework with 10K, 50K, and 100K subset of clean data from CC3M that constitute $\sim 0.3\%, 1.6\%, 3.3\%$ of the total pretraining dataset size, respectively. We present our results across the range of backdoor attacks in Table \ref{exp_table:ablation_selfi_training_size}.

\begin{table*}[h]
\begin{center}
\caption{Variation in attack success rate (ASR) and clean accuracy (CA) with finetuning dataset size in the CleanCLIP framework. All models were pretrained on CC3M with 1500 samples backdoored using the BadNet attack. All values are indicated in $\%$.}
\label{exp_table:ablation_selfi_training_size}
\resizebox{\textwidth}{!}{%
\begin{tabular}{lcccccc}
\hline
                 & \multicolumn{2}{c}{CleanCLIP-Unsup (CC10K)} & \multicolumn{2}{c}{CleanCLIP-Unsup (CC50K)} & \multicolumn{2}{c}{CleanCLIP-Unsup (CC100K)} \\\hline
\textbf{Attack Type}                & \textbf{CA} ($\uparrow$)       & \textbf{ASR} ($\downarrow$)      & \textbf{CA} ($\uparrow$)     & \textbf{ASR} ($\downarrow$)     & \textbf{CA} ($\uparrow$)     & \textbf{ASR}  ($\downarrow$)   \\\hline

\textbf{BadNet}      &   18.71 &  53.00 & 18.40   &  50.32 & 18.10 & \textbf{10.46}       \\
\textbf{Blended}               & 17.98  & 5.9  & 18.26  & \textbf{1.74} & 18.14 &   7.2\\
\textbf{WaNet}   &  18.18 &  0.16&   18.82 &  0.02  & 18.69 & 0.1\\
\textbf{Label Consistent}     & 18.95 & 27.52 & 18.82   &   20.28 & 18.99  &   11.08\\\hline
Average & 18.45 & 21.65 & \textbf{18.57} & 18.09 & 18.45 &\textbf{ 7.21}\\\hline
\end{tabular}%
}
\end{center}
\end{table*}

We find that finetuning on 10K data points leads to an average attack success rate of $21.65\%$ across the backdoor attacks which reduces to $7.21\%$ when the finetuning dataset size is increased to 100K. However, we find that the dependence of attack success rate on the finetuning dataset size is attack-specific. Specifically, the patch-based BadNet and Label-consistent trigger associations are not forgotten in the small data regime, whereas the non-patch-based Blended and WaNet triggers are much easier to forget with small data size. Overall, our results indicate that the visible patch-based attacks, although are easily detectable by humans, they are much difficult to forget by the model, in comparison to the invisible non-patch backdoor attacks. Additionally, we observe that the clean accuracy does not change much with the change in the finetuning dataset size.

\subsection{Effect of Supervised Finetuning Dataset Size}
\label{exp:sufi_dataset_size}

While performing supervised finetuning on a target dataset, here, we investigate the effect of varying the amount of labeled data on the clean accuracy and the attack success rate. To do so, the poisoned CLIP vision encoder is finetuned with 5K, 10K, and 50K images from the ImageNet-1K training data. We make sure that each class contains an equal number of images. We present our results across the range of backdoor attacks in Table \ref{exp_table:ablation_sufi_training_size}.

\begin{table*}[h]
\begin{center}
\caption{Variation in attack success rate (ASR) and clean accuracy (CA) with finetuning dataset size in the supervised finetuning framework. All models were pretrained on CC3M with 1500 samples backdoored using the BadNet attack. All values are indicated in $\%$.}
\label{exp_table:ablation_sufi_training_size}
\resizebox{\textwidth}{!}{%
\begin{tabular}{lcccccc}
\hline 
& \multicolumn{2}{c}{Sup. Finetuning (5K)} & \multicolumn{2}{c}{Sup. Finetuning (10K)} &
                 \multicolumn{2}{c}{Sup. Finetuning (50K)} \\\hline
\textbf{Attack Type}               & \textbf{CA} ($\uparrow$)       & \textbf{ASR} ($\downarrow$)       & \textbf{CA} ($\uparrow$)      & \textbf{ASR} ($\downarrow$)      & \textbf{CA} ($\uparrow$)     & \textbf{ASR}  ($\downarrow$)   \\\hline

\textbf{BadNet}       &    12.43&   0&21.88 &  0 &  40.86 & 0        \\
\textbf{Blended}             &   12.88&   0&21.82  &  0&  41.34 & 0  \\
\textbf{WaNet}                  & 12.81 & 0 &  21.86  &   0 & 40.43 & 0\\
\textbf{Label Consistent}     & 12.7 & 0 & 21.85   & 0   & 41.42 & 0.17  \\\hline
Average & 12.7 & 0 & 21.85 & 0 & 41.01 & 0  \\\hline
\end{tabular}%
}
\end{center}
\end{table*}

Unsurprisingly, we find that increasing the amount of labeled data for supervised finetuning monotonically increases the clean accuracy on the ImageNet-1K validation set i.e., it increases from $\sim 13\%$ to $\sim 41\%$ as the data increases from 5K to 50K. However, we find that the attack success rate is $\sim 0\%$ oblivious to the amount of finetuning dataset, across the backdoor attacks. This might be attributed to the catastrophic forgetting of the pretrained representations even at the small data scale while finetuning.

\section{Related Work}
\label{relatedwork}

\subsection{Multimodal Contrastive Learning}
Contrastive Learning \cite{chopra2005learning, hadsell2006dimensionality} was originally developed to learn self-supervised representations from individual modalities. The technique brings together the representations of an augmented version of a sample with the original sample while separating the representations of random samples in the embedding space. Recently, this method has been extended to the multimodal context, specifically for paired image-text data. Multimodal contrastive models such as CLIP \cite{radford2019language}, ALIGN \cite{ALIGN}, and BASIC \cite{pham2021combined} have been trained on large-scale data scraped from the web. This has led to their unprecedented ability to perform zero-shot classification and their robustness to distribution shifts. Several works have further extended this approach using visual self-supervision \cite{mu2022slip}, nearest-neighbor supervision \cite{li2021supervision}, consistency regularization \cite{goel2022cyclip}, as well as adding additional multimodal knowledge to the training process \cite{zellers2021merlot, zhang2021vinvl, desai2021virtex, li2022blip,alayrac2022flamingo}. Previous studies \cite{mu2022slip,li2021supervision} have combined self-supervised learning with CLIP pretraining to learn better visual representations. Related to our work, a concurrent work \cite{yu2023mitigating} proposes a novel approach that addresses spurious correlations during fine-tuning by leveraging a multi-modal contrastive loss function to explicitly separate spurious attributes from the affected class. However, we motivate the need for self-supervised learning with multimodal contrastive learning to encourage the model to learn representations, independent of each other. As a result, we could erase the spurious correlations learned by the CLIP model when exposed to the poisoned data.


\subsection{Backdoor Attack}

A backdoor attack is a data poisoning technique that involves embedding a hidden backdoor into examples in the training data of deep neural networks. The first instance of backdoor attacks for deep neural networks was presented by \cite{gu2017badnets}, where a small patch is embedded into an image, and its ground-truth class label is replaced with the target label in the training dataset. Since then, many backdoor attacks have been introduced that vary in terms of their imperceptibility and stealthiness. For example, \cite{chen2017targeted} introduced a blended key into images instead of a distinguishable patch in BadNet, while \cite{nguyen2021wanet} adopted image warping that deforms the image while preserving its content, and \cite{turner2019label} focused on the label-consistency of the backdoor attacks to make them harder to detect.

Initially, backdoor attacks were designed to attack neural networks that operate with unimodal data \cite{barni2019new,nguyen2020input,zeng2021rethinking,dai2019backdoor,chen2021badnl,jia2022badencoder,saha2022backdoor}. However, \cite{carlini2021poisoning} was the first to provide a framework to successfully attack multimodal contrastive models using the BadNet backdoor trigger, with just $0.01\%$ of the pretraining data. In this work, we find that (a) their framework applies equally well to various backdoor triggers, such as Blended, WaNet, and Label-consistent, and (b) we provide a defense mechanisms CleanCLIP, and demonstrate the effectiveness of supervised finetuning, to protect multimodal contrastive learning from these potent attacks.

\subsection{Backdoor Defense}

With the emergence of backdoor attacks, numerous studies have focused on identifying backdoor triggers in both the data and model, as well as removing backdoor triggers from the model itself \cite{wu2022backdoorbench}. Prior research such as \cite{dong2021black, chen2018detecting, tran2018spectral, qi2022fight, wang2019neural} has aimed to detect backdoor anomalies in input data and determine whether a model has been backdoored. In contrast, other studies \cite{wu2021adversarial, zeng2022adversarial, li2021neural, borgnia2021strong, du2019robust, li2021anti,yang2022not,liu2022friendly} have sought to minimize the harmful effects of backdoor triggers on models, with the goal of purifying them during training. Closely related to our work, \cite{huang2022backdoor} defend against backdoor attacks by employing self-supervised learning in their training process. Despite the success of these defense methods, they are tailored to backdoor attacks in the supervised learning paradigm, where there are limited number of classes bounded by the training dataset. However, multimodal contrastive learning at large-scale is applied in-the-wild and open-vocabulary. Hence, it is difficult to adapt the existing backdoor defenses, created for supervised learning, to the multimodal contrastive learning paradigm. In this study, we develop CleanCLIP, an unsupervised finetuning defense, and evaluate the effectiveness of supervised finetuning as strong candidates for backdoor defense in real-world use cases of the CLIP model. Additionally, we show that the multimodal adaptation of Anti-Backdoor learning \cite{li2021anti}, proposed for attacks in supervised learning, does not defend against backdoor attacks in CLIP.
\section{Conclusion}
\label{conclusion}

We presented CleanCLIP, a framework to defend against the potent backdoor attacks on multimodal contrastive pretraining, in CLIP. The main benefits of CleanCLIP stem from the balance between the self-supervised objective component that ensures that weakens the spurious associations learned by the individual modalities. As a result, it effectively reduces the ASR of a wide range of backdoor attacks. Furthermore, CleanCLIP does not make any assumptions regarding the target label, type, or amount of the backdoor attack for effective defense. Furthermore, we show that backdoor attacks lose their potency when the CLIP vision encoder is subjected to supervised finetuning using a labeled dataset. We believe this work can a valuable precedent for developing defenses against data poisoning attacks in multimodal contrastive learning. However, our work could also inspire further research on designing specialized multimodal backdoor attack strategies that remain potent under CleanCLIP defense.

\section{Acknowledgement}

This research is supported by a Sony Faculty Innovation Award, a CISCO Research Award, and a Sloan Fellowship. Hritik Bansal is supported in part by AFOSR MURI grant FA9550-22-1-0380. We also want to thank Da Yin, Ashima Suvarna, and Gantavya Bhatt for their helpful suggestions. 

\bibliography{main}
\bibliographystyle{plain}

\newpage

\appendix

\section{Backdoor Triggers Settings}
\label{app:backdoor_settings}


\begin{itemize}
\item For the BadNet attack, we add a $16\times16$ patch with each pixel sampled from a Normal distribution, $\mathcal{N}(0,1)$, to a random location in the image.
\item For the Blended attack, the poisoned image is obtained as $x' = 0.8 \times x + 0.2 \times n$, where $x$ is the clean image and $n$ is a noise tensor having the same shape as $x$ and containing uniform random values in the range $[0, 1)$.
\item For WaNet, we follow the setup used by \cite{qi2022fight} for ImageNet and use control grid size $k = 224$ and warping strength $s = 1$ and train models without the noise mode.
\item For the label-consistent attack, we sample images containing the target class label in the caption, and apply a trigger similar to the one used for BadNet while leaving the corresponding caption unchanged.
\end{itemize}   

\begin{figure*}[h]
    \centering
    \includegraphics[width=0.75\linewidth]{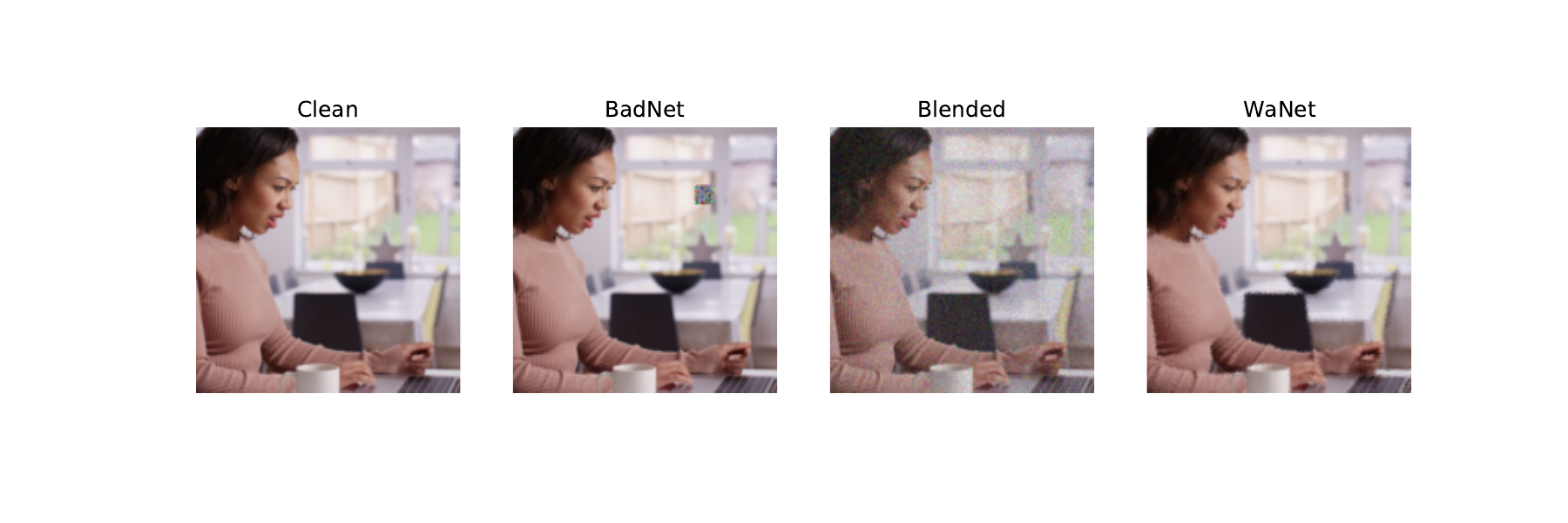}
    \caption{Examples of images poisoned using various backdoor attacks.}
    \label{fig:attack_examples}
\end{figure*}

\section{Cluster of the Target Class Images}

We find that the “clean” target class images lie in the cluster of the “clean” images in the embedding space for the poisoned model, and thus have a large distance from the backdoored images ($d=1.5$). After cleaning, the “clean” target class images lie very close to the “dirty” images in the image space ($d = 0.5$).

\begin{figure}[h]
    \centering
    \includegraphics[width=\linewidth]{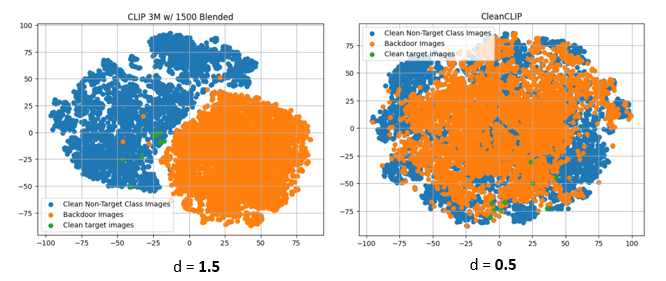}
    \caption{t-SNE plot of the image space.}
    \label{fig:my_label}
\end{figure}

\section{Does CleanCLIP work on Linear Probing?}

We train a linear classifier, using clean ImageNet-1K data, on a CLIP vision encoder learned by pretraining on CC3M w/ 1500 BadNet poisons. This model achieves an ASR of 89.81\%. However, the linear classifier on top of the CleanCLIP version of the poisoned model achieves an ASR of just 3.73\% without any reduction in the clean accuracy of 40\%. We observed similar behavior in other attacks.

\section{Baseline: Anti-Backdoor Learning (ABL) in Multimodal Contrastive Learning}
\label{appen:baseline_abl}

Since our defense strategies operate in the finetuning regime on the clean data, it is pertinent to benchmark their performance against strategies during the pretraining phase with the poisoned data. However, to the best of our knowledge, there has been no prior work to defend the models against the backdoor multimodal contrastive learning. Hence, as an additional contribution, we consider an adaption of the Anti-backdoor learning (ABL) \cite{li2021anti} framework, originally proposed for attacks in supervised learning, for multimodal contrastive learning.

Originally, ABL consists of two components -- (a) detecting backdoored samples from the pretraining data, followed by (b) the use of an additional objective that encourages the loss to maximize, instead of minimize, on the detected backdoored examples. In our adaptation to multimodal contrastive learning, we make use of a key insight that a \textit{clean} pretrained CLIP model would be unaware of the artificial associations between the backdoor trigger and the target label. Hence, the cosine similarity of the embeddings of a poisoned image and the caption containing the target label for a clean model would be low. Concretely, we compute the embeddings for all paired samples in the poisoned pretrained data using a pretrained CLIP from \cite{radford2021learning}. Subsequently, as a detection strategy we consider the $k$ samples with the lowest cosine similarities as poisoned. 

We denote the set of these $k$ samples as $\tilde{\mathcal{D}}_p$ and the remaining samples as $\tilde{\mathcal{D}}_c$, $\mathcal{D} = \tilde{\mathcal{D}}_p \cup \tilde{\mathcal{D}}_c$. Finally, we unlearn the detected backdoor examples by introducing an additional constraint to reduce the cosine similarity between the paired image and text representations of the samples in $\tilde{\mathcal{D}}_p$ to 0.  Formally, the ABL loss during pretraining looks like:

\begin{equation*}
    \mathcal{L}_{\text{ABL}} = \mathcal{L}_{\text{CLIP}}(\tilde{\mathcal{D}}_c) + \alpha \cdot \frac{1}{|\tilde{\mathcal{D}}_p|} \sum_{\tilde{\mathcal{D}}_p} [\langle I_i^e, T_i^e\rangle^2]
\end{equation*}
where $\mathcal{L}_{CLIP}$ is the CLIP training objective (Eq. \ref{eq:clip_final}) and $\alpha$ is a hyperparameter that controls the relative strength of unlearning. For our experiments, we use $k = 10,000$ as the size of $\tilde{\mathcal{D}}_p$.

\section{Training Dynamics}

\subsection{How do the training dynamics of the backdoored and the clean examples vary during CLIP pretraining?}

\begin{figure*}[h]
\centering
\subfloat[\centering \small{CLIPScores during training under BadNet attack.}]{{\includegraphics[width=0.495\linewidth]{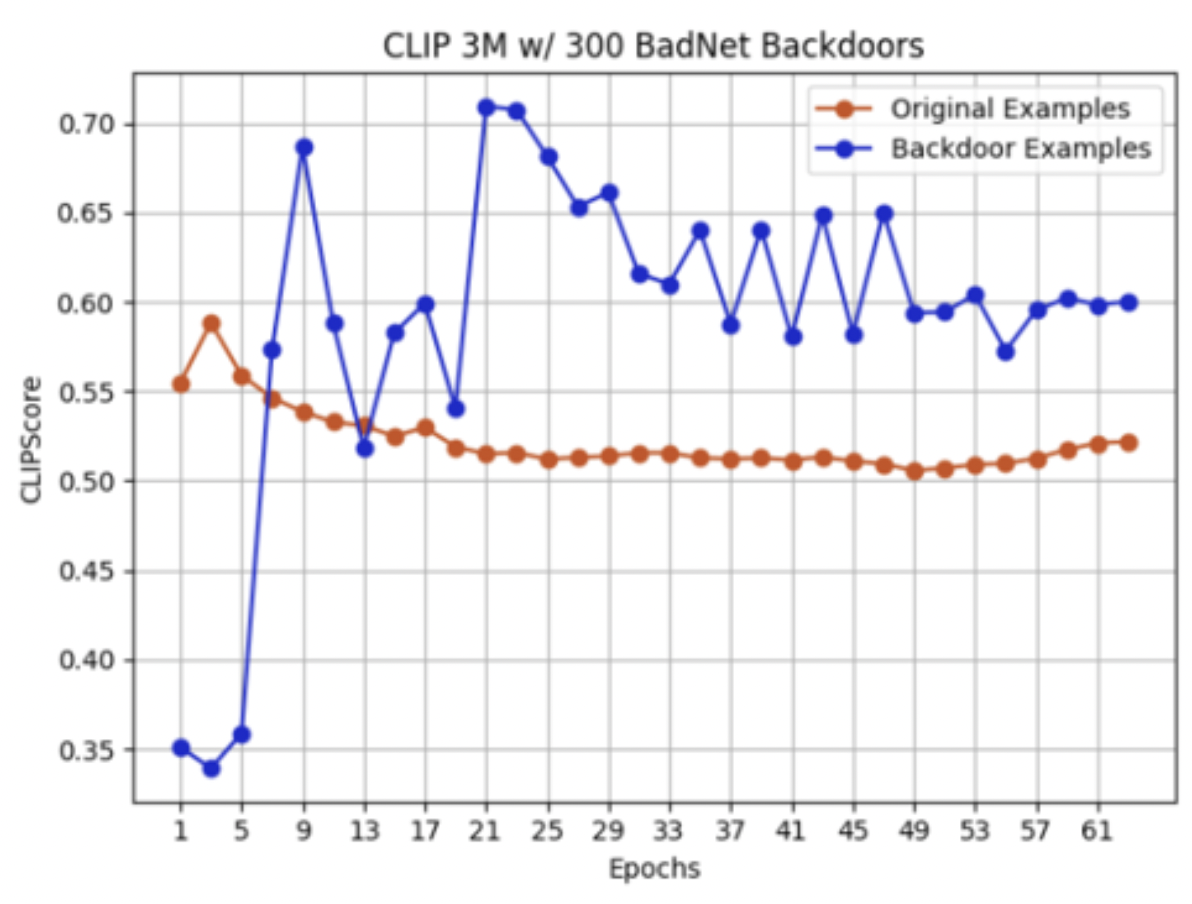}}}
\subfloat[\centering \small{CLIPScores during training under Blended attack.}]{{\includegraphics[width=0.5\linewidth]{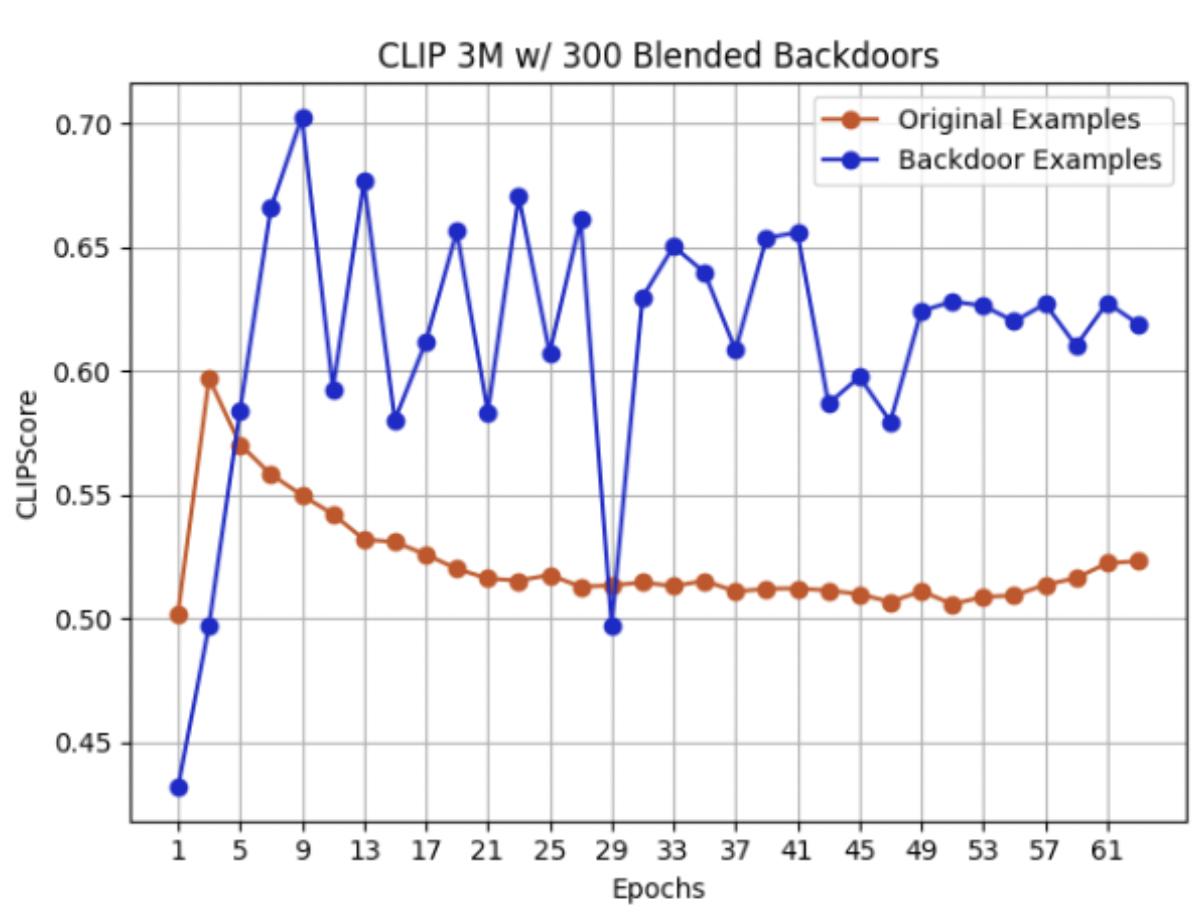}}}
\caption{Variation in the cosine similarity between embeddings of images and their corresponding texts (referred to as \textit{CLIPScore} for original (clean) and backdoored images during training. It can be seen that the CLIPScores of backdoored samples increase much more quickly as compared to the original samples. The models in both plots were trained on CC3M with 300 poisoned samples. The plot for `original' images was approximated by averaging the CLIPScores of 10,000 images randomly sampled from the training set of CC3M. We observed similar trends in the case of 1500 poisoned samples in the pretraining data.}
\label{fig:app_training_dynamics}
\end{figure*}

We analyze the training dynamics of the clean examples and the poisoned examples when a CLIP model is pretrained on the poisoned data, as in \S \ref{exp:effofdefense}. We find that the CLIPScore \cite{hessel-etal-2021-clipscore} i.e., the cosine similarity between the representations of the image and its corresponding text, increases much rapidly for the poisoned images than the clean images (Figure \ref{fig:app_training_dynamics}). This indicates that the spurious correlations between the image and text, from the poisoned example, are learned early in the training phase. 

\subsection{Can we use the apparent difference in the backdoor dynamics for effective detection during pretraining itself?}

\begin{figure*}[h]
\centering
\subfloat[\centering\label{app:det1} \small{1.5M training data w/ 75 Blended poisons}]{{\includegraphics[width=0.3\linewidth]{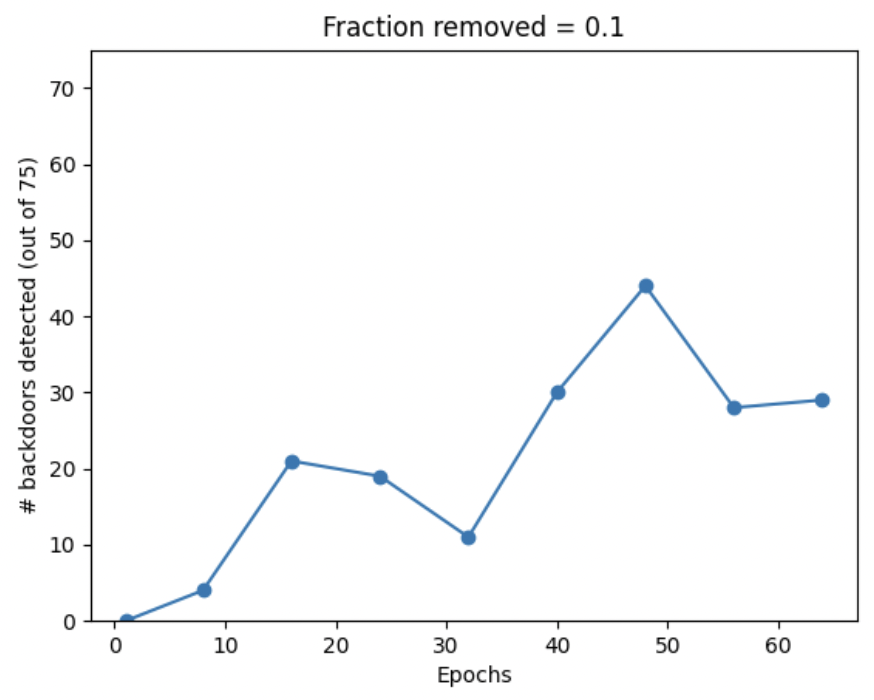}}}
\subfloat[\centering\label{app:det2} \small{1.5M training data w/ 300 Blended poisons}]{{\includegraphics[width=0.3\linewidth]{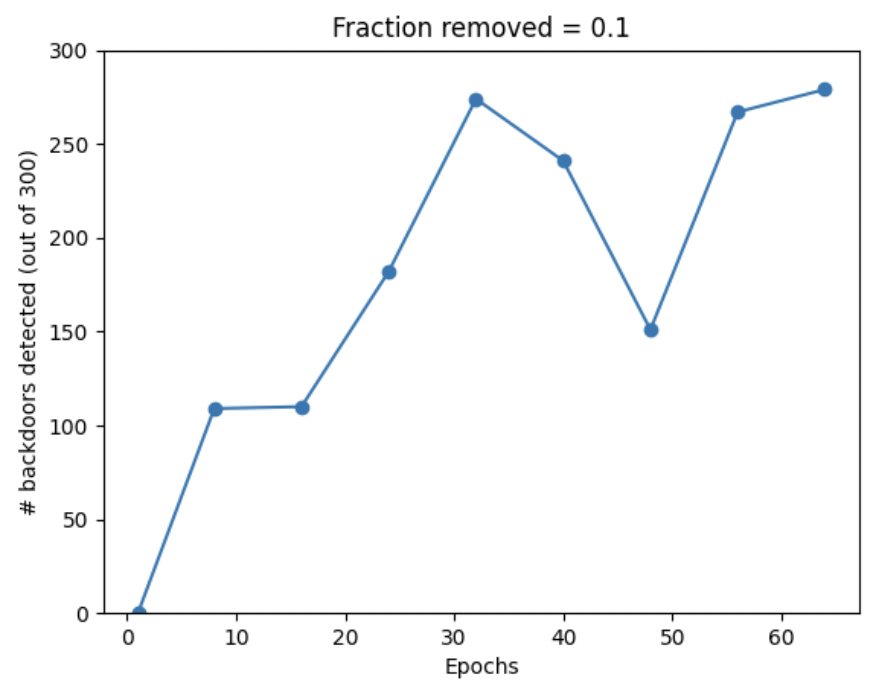}}}
\subfloat[\centering \label{app:det3} \small{1.5M training data w/ 1000 Blended poisons}]{{\includegraphics[width=0.31\linewidth]{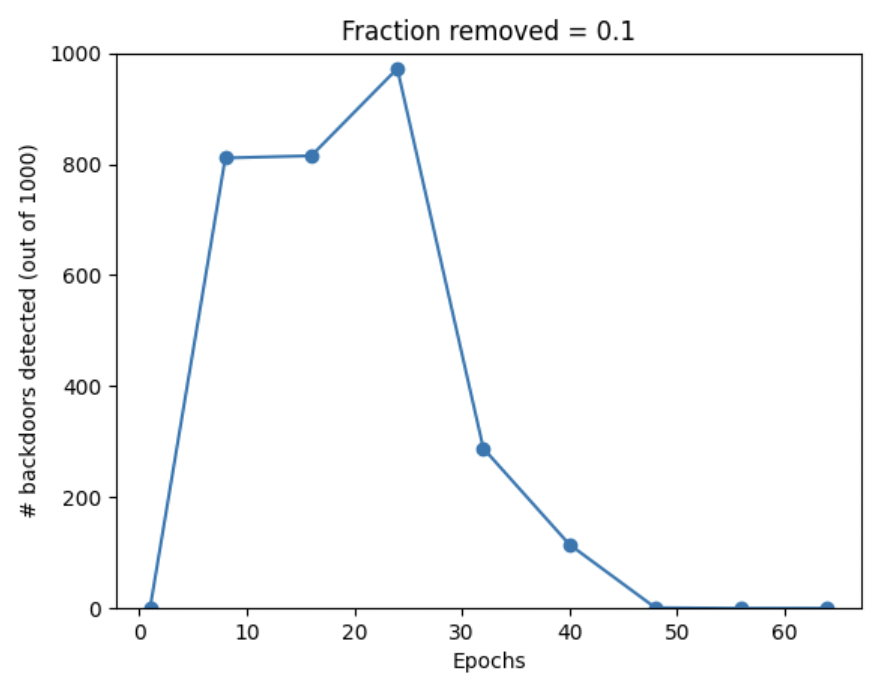}}}

\caption{Results of the strategy that aims to detect poisoned data during pretraining using the training dynamics of clean and poisoned samples. We pretrain CLIP on 1.5M samples from the CC3M training data attacked by the Blended attack with (a) 75, (b) 300, and (c) 1000 poisoned samples, respectively. Subsequently, we consider the top 10\% training samples, with the highest CLIPScore, at a given pretraining epoch as poisoned. We evaluate this strategy at various epochs during pretraining and find that there is no single epoch that works well across all settings.}
\label{fig:app_detection_results}
\end{figure*}

Since, we observe a clear distinction between the training dynamics of the clean and backdoor examples, it is imperative to study whether it is easier to detect the backdoored examples well before the pretraining ends. To that end, we consider $k$ samples with the highest cosine similarities at epoch $T$ as the potentially poisoned examples. We report the number of true positives i.e., the number of true backdoored examples that are captured in the $k$ detected examples in Figure \ref{fig:app_detection_results}. We show the results for a model trained on 1.5M data with Blended attack for various values of the detection epoch $T$. We find that the number of backdoors detected by the strategy can be sensitive to the choice of the particular epoch. For instance, we observe that the number of detections suddenly drops at Epoch 50 when we use $k = 0.1 |\mathcal{D}|$ where $|\mathcal{D}|$ is the size of the training data, in Figure \ref{app:det2}. We also find large qualitative variation in the results across the three models trained with 75, 300, and 1000 poisons, respectively. For instance, later epochs work well for the model trained with 75 poisons but not for the model trained with 1000 poisons.

\subsection{Can we use a set of the correctly detected poisoned examples to erase the impact of the backdoor trigger?}

In \S \ref{appen:baseline_abl}, we had used a CLIP model that is pretrained with 400M data, however, it is unaware of the characteristics of any specific backdoor attacks since it is not trained on them. To that end, we evaluate whether the detections from a CLIP model that is pretrained on the poisoned data be more useful to construct a stronger defense. Concretely, we considered the top 5,000 samples with the highest CLIPScore at epoch 8, chosen randomly, as backdoored samples and performed our adaptation of anti-backdoor learning. We find that even the unlearning objective failed to defend the model, since the undetected backdoor examples were enough to poison the model via multimodal contrastive loss. For instance, in the case of a CLIP model trained on 1.5M data with 1000 samples poised with the Blended attack, only 368 poisoned samples were correctly detected as backdoors, and the remaining undetected backdoor examples were enough to maintain the ASR to 98.53\%. Similarly, for the WaNet attack with 1000 backdoored samples out of 1.5M training samples, only 168 samples were detected and the ASR was 99.35\%. The potency of the backdoor attack remained high in our experiments even when the weight of the unlearning term was increased. We believe that exploring different detection and unlearning strategies that can effectively eliminate backdoor attacks during pretraining is an interesting direction for future work.

\end{document}